\documentclass[letterpaper, 10 pt, journal, twoside]{IEEEtran}
\usepackage{graphicx}
\usepackage{cite}
\usepackage{xcolor}
\usepackage[linesnumbered,ruled,vlined]{algorithm2e}
\usepackage{floatrow}
\usepackage{multirow}
\usepackage{makecell}
\usepackage{booktabs}
\usepackage{siunitx}
\usepackage[font={small}]{caption}
\DeclareCaptionFont{xipt}{\fontsize{11}{13}\mdseries}
\SetKwInput{KwInput}{Input}                
\SetKwInput{KwOutput}{Output}              

\IEEEoverridecommandlockouts                              
\usepackage{tikz}
\usepackage{textcomp}
\usepackage{hyperref}

\newcommand\copyrighttext{%
  \footnotesize \textcopyright 2020 IEEE. Personal use of this material is permitted.
  Permission from IEEE must be obtained for all other uses, in any current or future
  media, including reprinting/republishing this material for advertising or promotional
  purposes, creating new collective works, for resale or redistribution to servers or
  lists, or reuse of any copyrighted component of this work in other works.
  DOI:{10.1109/LRA.2020.3003886}}
\newcommand\copyrightnotice{%
\begin{tikzpicture}[remember picture,overlay]
\node[anchor=south,yshift=5pt] at (current page.south) {\fbox{\parbox{\dimexpr\textwidth-\fboxsep-\fboxrule\relax}{\copyrighttext}}};
\end{tikzpicture}%
}

\begin{document}

\title{Coverage Path Planning with Track Spacing Adaptation for Autonomous Underwater Vehicles
}

\author{Veronika Yordanova$^{1}$ and Bart Gips$^{1}$

\thanks{Manuscript received: February, 20, 2020; Revised May, 12, 2020; Accepted June, 05, 2020.} 
\thanks{This paper was recommended for publication by Editor Jonathan Roberts upon evaluation of the Associate Editor and Reviewers' comments.
This work was supported by NATO Allied Command Transformation (ACT).} 
\thanks{$^{1}$Veronika Yordanova and Bart Gips are with NATO STO Centre for Maritime Research and Experimentation (CMRE), La Spezia, Italy
        {\tt\small \{veronika.yordanova, bart.gips\}@cmre.nato.int}}%
}

\markboth{IEEE Robotics and Automation Letters. Preprint Version. Accepted
June, 2020}
{Yordanova \MakeLowercase{\textit{et al.}}: Track Spacing Adaptation for AUVs}

\maketitle

\copyrightnotice

\begin{abstract}

In this paper we address the mine countermeasures (MCM) search problem for an autonomous underwater vehicle (AUV) surveying the seabed using a side-looking sonar.
We propose a coverage path planning method that adapts the AUV track spacing with the objective of collecting better data.
We achieve this by shifting the coverage overlap at the tail of the sensor range where the lowest data quality is expected.
To assess the algorithm, we collected data from three at-sea experiments.
The adaptive survey allowed the AUV to recover from a situation where the sensor range was overestimated and resulted in reducing area coverage gaps.
In another experiment, the adaptive survey showed a 4.2\% improvement in data quality for nearly 30\% of the `worst' data.

\end{abstract}

\begin{IEEEkeywords}
Marine Robotics, Robotics in Hazardous Fields, Search and Rescue Robots, Motion and Path Planning, Reactive and Sensor-Based Planning
\end{IEEEkeywords}

\section{INTRODUCTION}

\IEEEPARstart{C}{overage} path planning (CPP) is the problem of defining a path, such that the sensor swath passes over all points in a given area.

Typical applications for coverage planning are surveillance, lawn mowing, vacuum cleaning, farming, painting and manufacturing.
It is hard to define the optimal coverage problem with a general set of requirements and limitations: most instances are NP-hard (non-deterministic polynomial-time hard), even the simplified ones \cite{arkin2000approximation}.
In this paper we address the mine countermeasures (MCM) search problem for an autonomous underwater vehicle (AUV) surveying the seabed for objects using a side-looking sonar.

\subsection{Mine countermeasures application}
\begin{figure}[htbp]
	  \centerline{\includegraphics[width=1\linewidth]{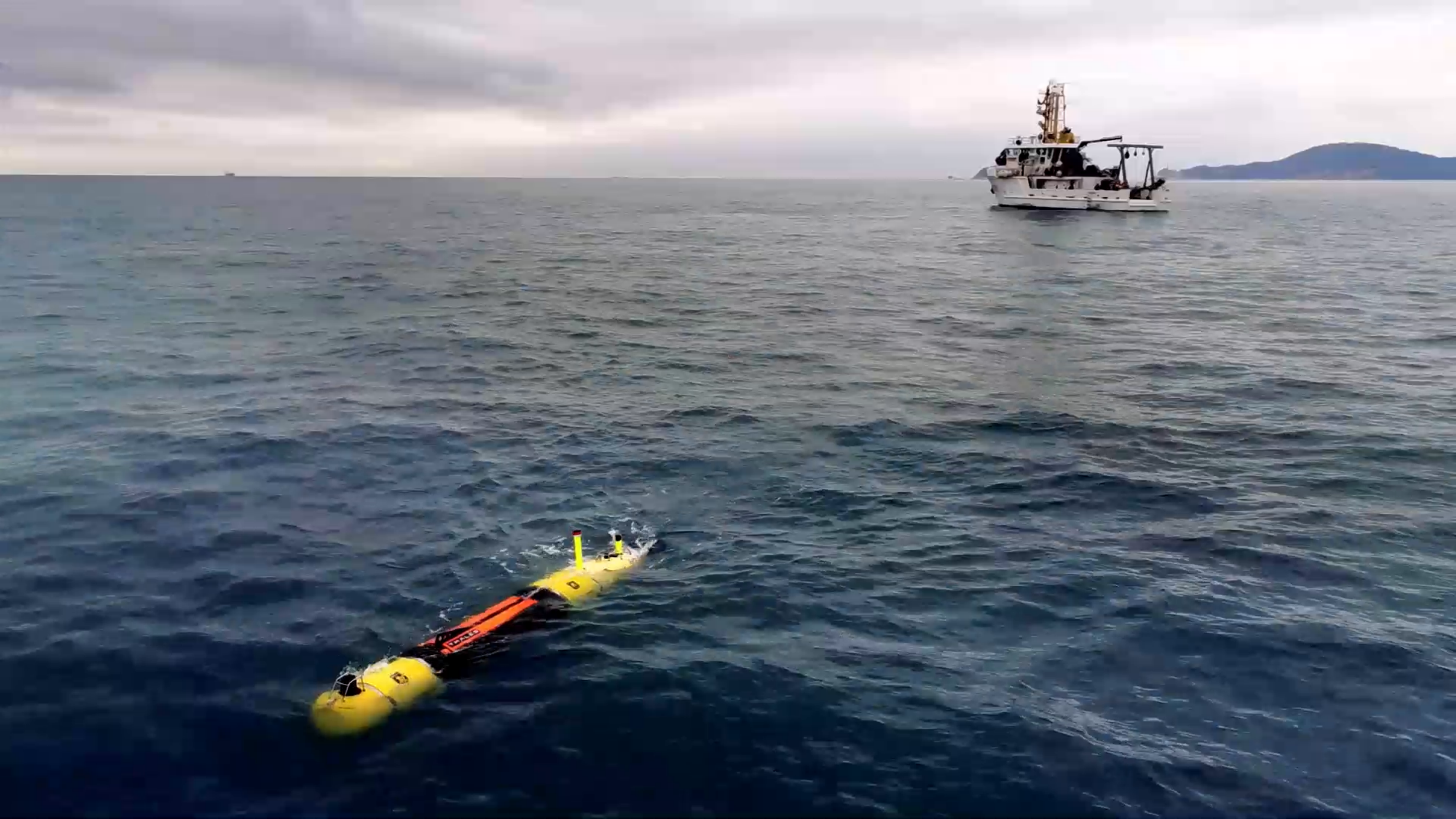}}
	  \caption{Deployed autonomous underwater vehicle (MUSCLE) from the coastal research vessel Leonardo in Liguria, Italy.}
	  \label{fig:muscle}
	\end{figure}

Naval mines are a threat for military and civilian operations.
The process of mine clearing starts with surveying an area for mine-like objects using a sonar sensor.
The detections need to be further classified and then the type of the mine identified, before it is neutralised.
Currently, large manned minehunters and minesweepers perform these tasks, however, recent research suggests that AUVs could replace them and make MCM operations safer, more efficient and economic \cite{dugelay2019enabling}.
Figure \ref{fig:muscle} is an example of an AUV deployed from a ship at the start of an MCM experimental trial.

One of the big open research questions in the field is how to improve the probability of target detection, $P_d$.
To address this problem, we propose an approach that aids the sonar data collection by connecting through-the-sensor (TTS) performance estimation and real-time vehicle path replanning.
We focus on the search phase, as an instance of a CPP problem, and consider the remaining MCM phases outside the scope of this paper.

\subsection{Related work}

A survey paper on CPP, published in 2001 \cite{choset2001coverage}, classified existing methods, in an attempt to generalise the problem; a more recent survey is available from 2013 \cite{galceran2013survey}.
Partitioning the area and turn-minimisation strategies are common ways to improve efficiency in robotic CPP applications \cite{huang2001optimal, bochkarev2016minimizing, vandermeulen2019turn}.
A CPP algorithm for an AUV adapting to an \textit{a priori} known bathymetry map explored specifics in the marine robotics domain \cite{galceran2013planning}.
Successful data collection strategies countering ripples \cite{williams2010sand} and currents \cite{williams2016adaptive} have been demonstrated at sea.
Recently, we combined data-driven and efficiency metrics in an AUV track orientation approach in order to adapt to a sand ripple seabed and minimise the number of turns performed by the AUV \cite{yordanova2019coverage}.

In addition to track angle adaptation, achieving a more consistent area coverage can be addressed by controlling the spacing between the tracks.
Compared with most other applications, where the CPP sensor swaths are constant and uniform \cite{hernandez2017auv}, the side-looking sonar, mounted on an AUV, has a non-uniform coverage and a nadir gap under the vehicle (see Figure \ref{fig:scheme_range}).

One of the first successful implementations of an AUV track spacing adaptation used image quality information as an input.
Image quality was measured by the coherence, or ping-to-ping correlation, of the synthetic aperture sonar (SAS) data.
A threshold, defining the admissible data was based on ``extensive experience visually assessing the quality of sonar images'' \cite{williams2012auv} and verified in a sensitivity analysis that assessed detection performance as a function of admissible coherence \cite{williams2012auv}, \cite{williams2015fast}.
A more recent approach of an AUV adapting its tracks using the coherence in SAS as a metric for quality is described in \cite{Geilhufe2018}.

In both track spacing adaptation examples, if a coverage gap is left due to a mismatch between expected and real sensor quality, the vehicle passes over that space again with its next track.
Reacting only to a single track, rather than considering the coverage of the survey area as a whole, can lead to suboptimal resource allocation, such as the need of additional tracks at the end of the region.

\section{Problem Definition}
To study the mine countermeasures coverage path planning problem, we define the following objectives:
	\begin{itemize}
	\item Given an area (convex, simple polygon) and a sensor swath, defined by range ($r_{max}$) and nadir gap ($r_{min}$), generate a lawnmower path that covers the mission area. In addition, when regions get ensonified more than once, this coverage overlap needs to be pushed to the areas where the $P_d$ is lower.
	\item Given data available from the sensor, adapt the track spacing based on data quality measures.
	\end{itemize}

The algorithm needs to take into account two distinct inputs: \emph{operational} and \emph{data-related}. 
The operational inputs are parameters controlled by operators or restricted by the available equipment, and consist of mission area and sensor range.
The data-related inputs are environmental variables that are not controlled.
The resulting track spacing output should be based on a tradeoff between overall coverage, number of tracks and data quality.

\subsection{Assumptions}

During an MCM seabed survey mission, the issue of obstacle avoidance may generally be ignored.
This reduces the problem to taking into account only the boundaries of an area when optimising the track planning.

In this paper, we bound our experiments to a rectangular box, instead of using a more general polygon shape.
In our previous work, we addressed coverage in a convex polygon area using angle adaptation \cite{yordanova2019coverage}, but decoupled track angle from spacing for these experiments, as the variables are independent.

We assume a flat and horizontal seabed.
Regardless, if the survey area is sloped, we consider that adapting the angle of the vehicle's tracks would result in a better and more uniform coverage \cite{galceran2013planning}.
When there is clutter, or objects that can obstruct the sonar, this is reflected in the coverage quality.
Sometimes the area is unhuntable, other times a second look from a different angle is required.
We consider these problems to be orthogonal to optimizing track spacing, and should therefore be considered separately.

We assume the vehicle can exit the region that needs to be covered, where it can turn.
While this might not always be possible, further research in waterspace management is required to define the guiding operational constraints \cite{ferreira2019underwater}.

We have also made assumptions based on the vehicle's instrumentation.
The vehicle can keep at constant altitude with the help of a conductivity, temperature and depth (CTD) sensor and a Doppler velocity log (DVL).
The Inertial Navigation System (INS) provides small enough drift to make it negligible for the experiments in this paper.
For longer missions, an external navigation system can be assumed, as it can be available from a ship in the deployment area.

\subsection{Contribution}
In this paper, we present a coverage planner that adapts the track spacing of an autonomous underwater vehicle for mine countermeasures survey phase.
We use a heuristic method that optimises the AUV's full path plan, with respect to mission efficiency and data quality.
The path can be adapted on-the-fly based on the latest information, such as changes in sensor measured performance.
We deployed the algorithm on an AUV and collected experimental data comparing adaptive and predefined survey results.

\section{METHODS}
\label{methods_section}

The adaptive track spacing algorithm we propose takes into account the shape and size of a side-looking sonar sensor swath.
Our performance metric to separate `good' from `bad' data is based on an estimate of the sensor's ability to detect a target as a function of range.
We want to place the tracks of the AUV path in such a way, that if we need to ensonify a part of the area more than once, this should coincide with the area with the `worst' data.

\subsection{Sensor range}
\label{sensor_range}

\begin{figure}[htbp]
	  \centerline{\includegraphics[width=1\linewidth]{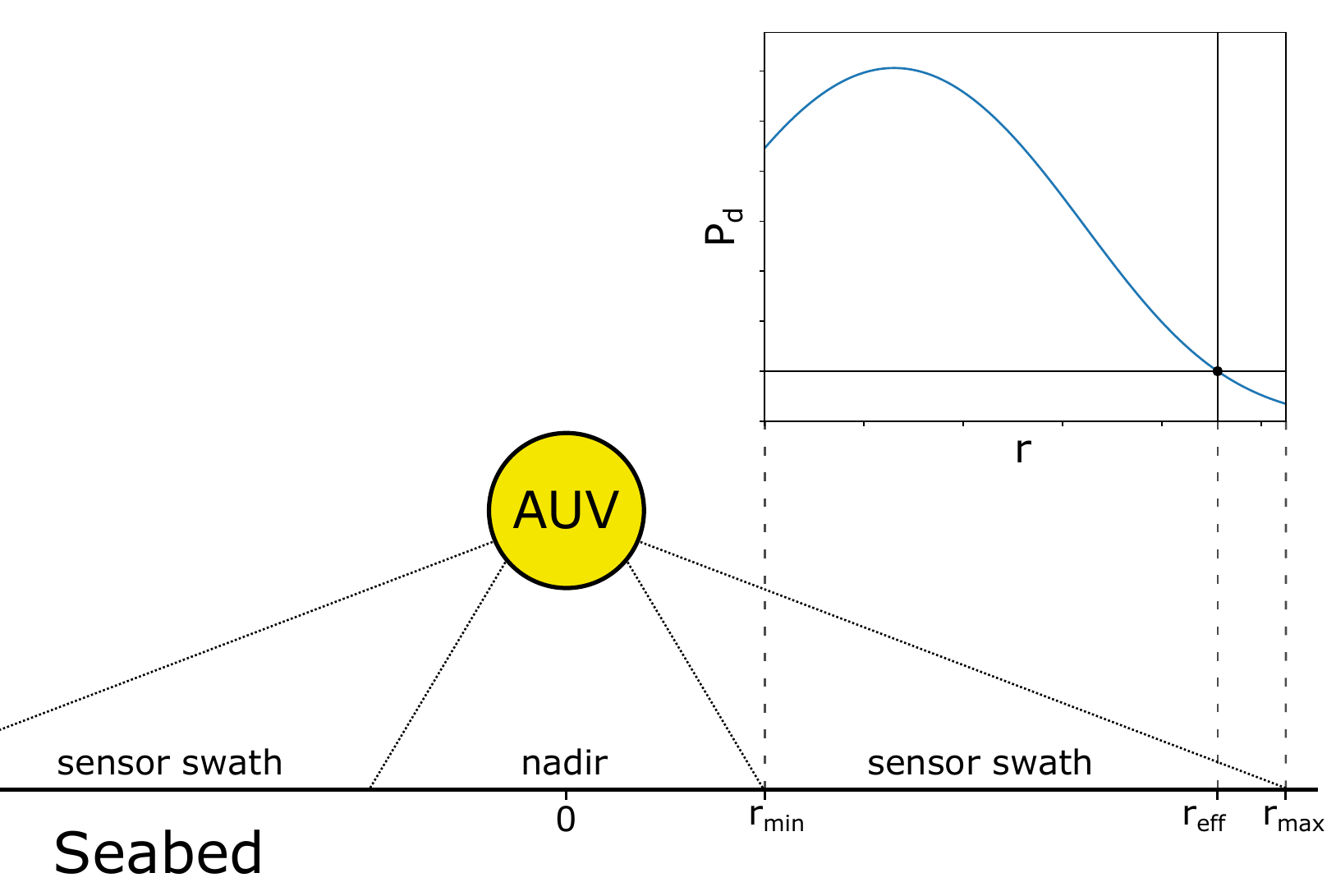}}
	  \caption{Schematic of the MUSCLE sensor range and nadir gap. In the top right we show the $P_d$ curve that determines $r_{eff}$, see Figure \ref{fig:pd_profile}.}
	  \label{fig:scheme_range}
	\end{figure}

For our experiments, we use the Mine-hunting UUV for Shallow-water Covert Littoral Expeditions (MUSCLE) --- a Bluefin-21 autonomous underwater vehicle, equipped with a side-looking synthetic aperture sonar (SAS) \cite{Baralli2013}.
Figure \ref{fig:scheme_range} shows a schematic of the MUSCLE sensor range ($r_{max}$) and nadir gap range ($r_{min}$).
The data quality deteriorates with range so the actual admissible data limit is $r_{eff}$.
It is computed by taking the intersection of the sensor range profile, given by a $P_d$ curve (explained in section \ref{sec:rrm}), and a threshold separating admissible from inadmissible data.
Setting this threshold is out of the scope of this paper.
Its limits can be set manually, when driven from operational requirements, or automatically, to match the variable sensor performance.
Changing this threshold will affect $r_{eff}$ and will therefore alter the resulting track spacing, but it will not fundamentally alter the workings our algorithm.

If we want to ensonify the nadir gap, we need to use paired tracks, where one pass ensonifies the area missed by the other.
This gives us a constraint on $r_{max}$ relative to $r_{min}$:

	\begin{equation}
	r_{max} \geq 3\times r_{min}.
	\label{eq:sensor_range_nadir}
	\end{equation}

The nadir gap width  ($r_{min}$) for the SAS on the MUSCLE was 40 meters per side, a result of the chosen sonar tilt angle and mode, and the altitude of the vehicle.
Equation \ref{eq:sensor_range_nadir} gives the minimum admissible value for the sensor range --- $r_{max}=120$ metres.
We do not consider the case when $r_{eff}$ is smaller than this value as this would require a different track placement model.

Based on operators' experience $r_{max}$ is not expected to exceed 150 metres.
Hence, for our adaptive survey, we assume the sensor range should vary between 120 and 150 metres.

\subsection{$P_d$ curves}
\label{sec:rrm}
During the AUV survey, we wish to gather as much high-quality data as possible, but image quality generally degrades with range.
Therefore, our coverage path planning algorithm needs a performance metric that captures this range-dependence of the data quality. 
For example, Paull at al. \cite{paull2018probabilistic,paull2014area} have used the Extensible Performance and Evaluation Suite for Sonar (ESPRESSO) \cite{davies2006espresso} to generate a curve expressing the coverage as a function of lateral range.
In our case we shall use an explicit estimate of the probability that a target is detected ($P_d$) by the onboard Automatic Target Recognition (ATR) algrithm \cite{williams2015fast}.

This probability is estimated based on through-the-sensor (TTS) features such as image quality and environmental characteristics \cite{Gips2018a}.
This statistical model has been developed for the generation of residual risk maps (RRMs, e.g. Figure \ref{fig:all_exp_rrm_data}) that quantify the likelihood of overlooking a target (i.e. $1-P_d$). 
Therefore we can use this metric to estimate the performance of the AUV during or after a mine-hunting mission.

\begin{figure}[htbp]
	  \centerline{\includegraphics[width=.99\linewidth]{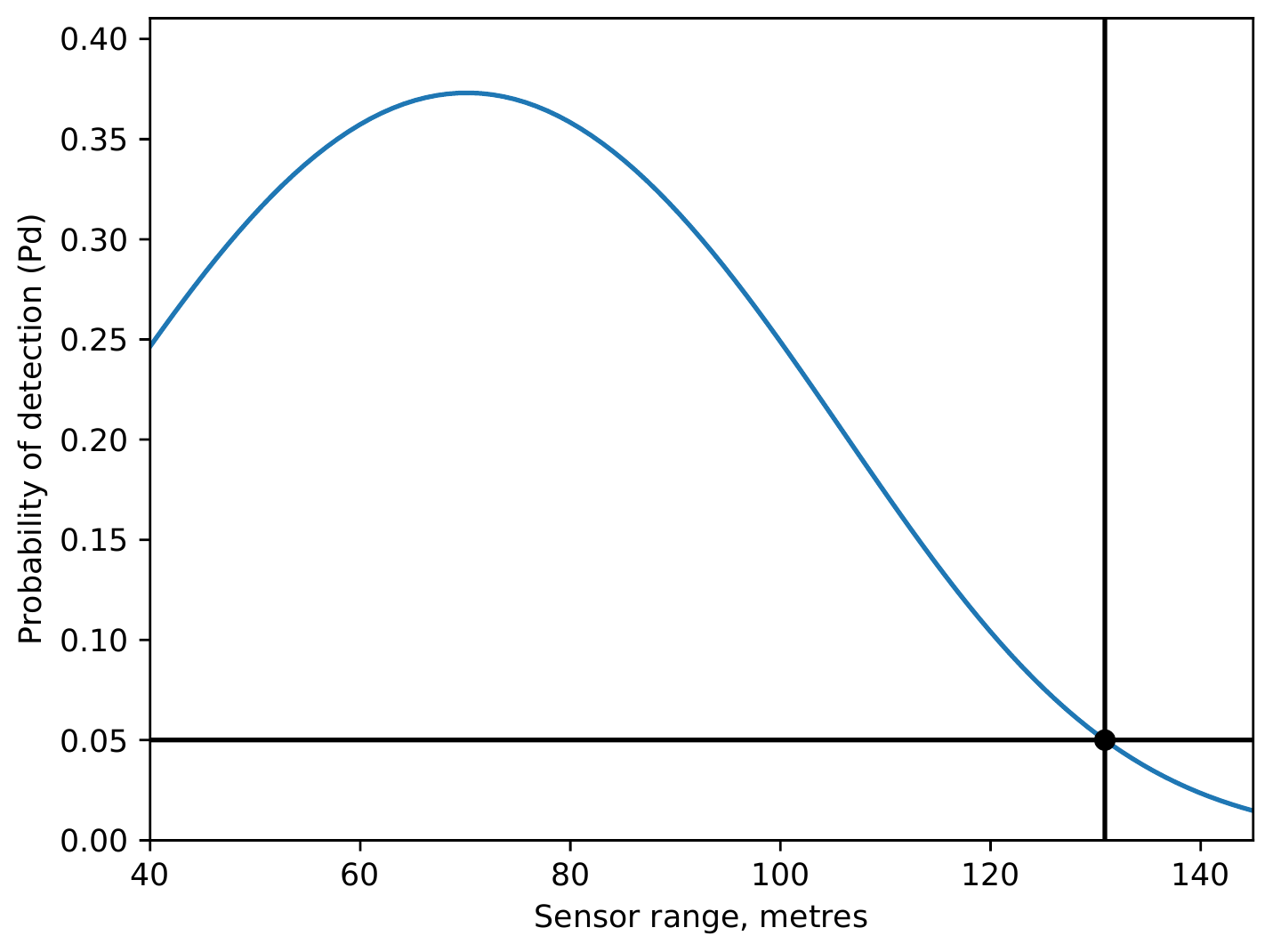}}
	  \caption{The expected value for the probability of detection ($P_d$) expressed as a function of lateral sensor range (also known as a $P(y)$ curve). Data from Experiment 1 in Section \ref{results} is used to generate this instance of $P_d$ curve. The black dot and vertical and horizontal lines indicate the threshold for data that is considered `good'.}
	  \label{fig:pd_profile}
	\end{figure}

In order for this model to be useful for our path planning algorithm, we extract an estimate of $P_d$ as a function of range alone, by integrating out all other TTS features based on SAS images collected in the preceding track (or averages based on previous missions for the first track). 
As such, the estimated performance as a function of range is updated after every track, leading to a possible change in $r_{eff}$.
An example of a resulting curve is shown in Figure \ref{fig:pd_profile}, which is also known as a $P(y)$ curve in the naval MCM community.
When the same piece of seabed is ensonified more than once, the coverage is added multiplicatively:

\begin{equation}
\label{eq:RRM}
 RR=\prod_i (1-{P_d}_i)
\end{equation}

where $RR$ is the residual risk intensity in a certain gridpoint (Figure \ref{fig:all_exp_rrm_data}), ${P_d}_i$ is the probability of detection at that gridpoint for look $i$, and $i$ iterates over all looks on this gridpoint.

The reduced probability of detection at both ends of the curve reflects the reduced ability of the vehicle to collect high quality data, with an optimum at a range around 70 m.

The threshold that determines the limit of admissible or `good' data is set at 0.05 and determines $r_{eff}$ as indicated in Figure  \ref{fig:scheme_range} and \ref{fig:pd_profile}.
At this stage, this is an arbitrary number and further research is needed to evaluate and adapt this threshold.
Importantly, the $P_d$ curve depends on elements such as the AUV, the sensor, and the ATR.
This could mean that Figures \ref{fig:pd_profile}, \ref{fig:all_exp_rrm_data} and \ref{fig:hist} may be completely different for different AUVs.
Regardless, any conclusions for our track adaptation methods presented here are equally valid for different $P_d$ curves or alternative metrics that quantify sensor performance as a function of range such as a $P(y)$ curve generated by ESPRESSO.

\subsection{Adaptive track spacing}

%
%



\begin{figure}[htbp]
	  \centerline{\includegraphics[width=1\linewidth]{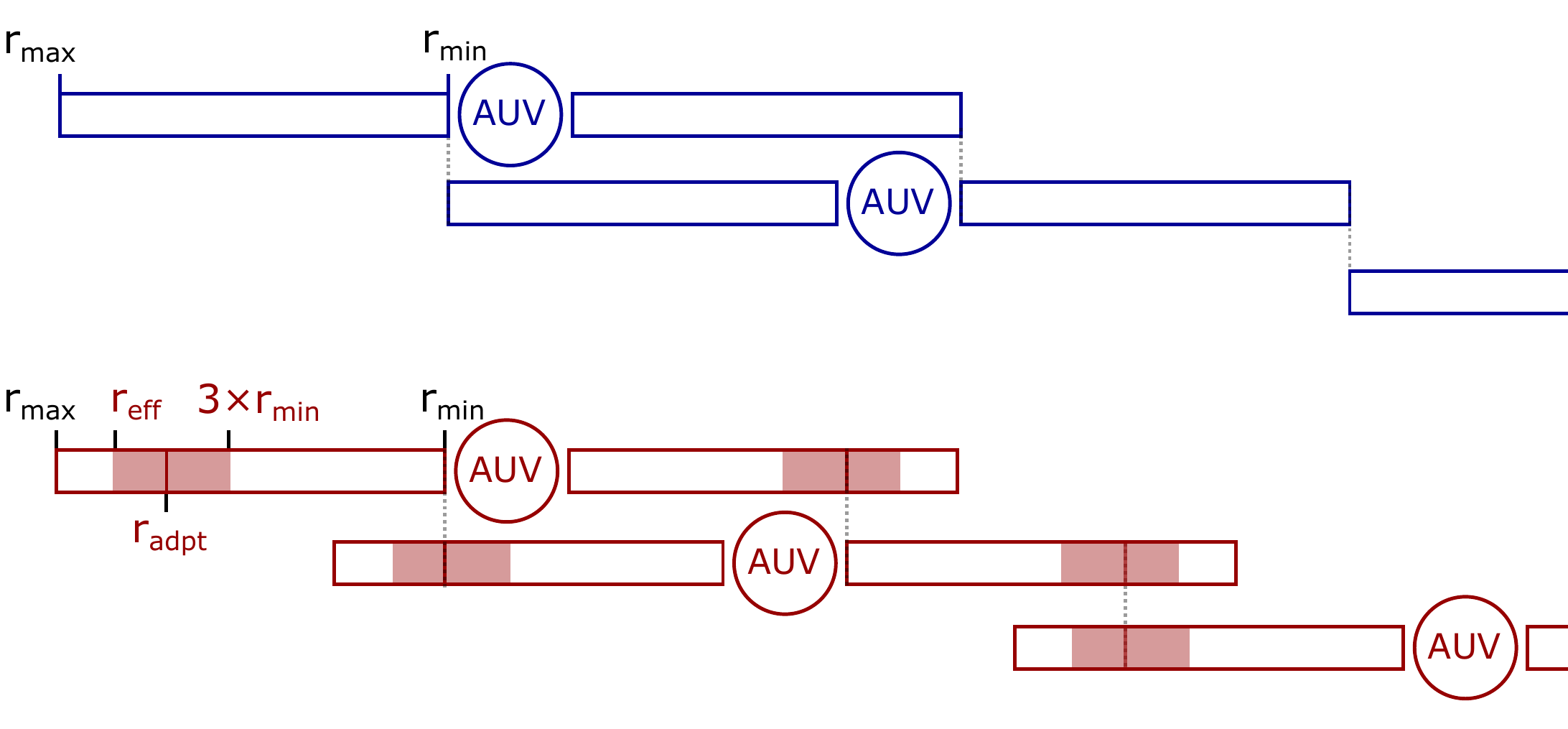}}
	  \caption{Schematic of AUV track alignment based on the adaptive track spacing algorithm. The boxes indicate the sonar sensor coverage. In the top, in blue, is an alignment following a predefined mission. The tracks are placed in pairs and spaced based on a predefined {$r_{max}$} and $r_{min}$  values. Within a pair, the maximum range of one track is aligned with the minimum of the other. In this way both nadir gaps are covered. Between tracks the maximum ranges are aligned to maximize coverage.
	In the bottom, in red, we illustrate the adaptive track spacing. In this case we adapt the virtual sensor range $r_{adpt}$. This can vary between  $3\times r_{min}$ (eqn \ref{eq:sensor_range_nadir}) and $r_{eff}$ (based on data) as indicated by the shaded area.
	Now $r_{adpt}$ takes over the role of $r_{max}$, leading to a tighter track spacing.}
	  \label{fig:range_to_spacing}
	\end{figure}

A straightforward way to achieve full area coverage is to place the AUV tracks in a sequence, depending on the shape and size of the sensor's swath (see the alignment in blue in Figure \ref{fig:range_to_spacing}).
The sonar sensor range is affected by the environment and it is often hard to estimate the range at which `good' data can be collected ($r_{eff}$).
This is why during operations, the most conservative sensor range is chosen in order to make sure there are no coverage gaps at the end of the mission.

Knowing the real sensor range, rather than choosing a conservative estimate, gives us the opportunity to define an AUV path that requires the least number of tracks in order to achieve full area coverage.
However, since the data quality produced by the sonar sensor is not uniform (Figure \ref{fig:pd_profile}), the placement of the tracks also matters, not just the number.

Figure \ref{fig:range_to_spacing} gives an illustration of a predefined and adaptive track placement strategies.
An example of track alignment based on predefined sensor range is shown in blue on Figure \ref{fig:range_to_spacing}; no adaptation is enabled and the tracks are positioned sequentially in pairs.
The overlap is close to the nadir gap, and none near the tail of the sensor range (see Figure \ref{fig:pd_profile}).

The red tracks show the adaptive track spacing.
It illustrates the different reference points we use to define the adaptive track spacing algorithm:
\begin{itemize}
\item $r_{max}$ --- outer sensor ensonification range
\item $r_{min}$ --- inner sensor ensonification range (nadir gap)
\item $r_{eff}$ --- estimated limit for `good' data collection
\item $3\times r_{min}$ --- paired-tracks limit, given by equation \ref{eq:sensor_range_nadir}
\item $r_{adpt}$ --- adapted sensor range --- can vary between $3\times r_{min}$ and $r_{eff}$
\end{itemize}

$r_{adpt}$ aligns with $r_{min}$ of the previous track to the left, and with the subsequent track's $r_{adpt}$ to the right.
This way we achieve coverage where the sensor range degrades --- at both ends of the $P_d$ curve (see Figure \ref{fig:pd_profile}).
Our aim is to optimise the choice of $r_{adpt}$ given the area size and the sensor range.

The Polygon Adaptation approach (Algorithm \ref{algo:poly_adpt}) relies on an exhaustive search of sequential track placement, given the width of the search area in the sweep direction, and the possible sensor ranges, $r\in[a,b]$.
The lower boundary of the interval is defined by the relationship in equation \ref{eq:sensor_range_nadir}.
The upper boundary is defined by the intersection of the $P_d$ curve with a threshold that separates admissible from non admissible data ($r_{eff}$), or some initial estimate for the sensor ensonification when there is no available data ($r_{max}$ in Figures \ref{fig:scheme_range} and \ref{fig:range_to_spacing}).
We limit the state space of the algorithm between these boundaries as they provide full coverage with no gaps, using only a pair of tracks to cover the nadir gap.
The state space is discretised at 1 metre intervals.

\begin{algorithm}
  
  \KwInput{Width of polygon in the sweep direction, $W$, Sensor range interval, $r \in [a,b]$}
  \KwOutput{Adapted sensor range, $r_{adpt}$}
  \For{all $r \in [a,b]$}
  {
	$n_t$ number of tracks for path within $W$
  }
  $r_{adpt} \gets $ smallest $r$ that gives min number of tracks $n_t$
  
  \Return{$r_{adpt}$}

\caption{Polygon adaptation for optimal coverage overlap.}
\label{algo:poly_adpt}
\end{algorithm}

The heuristic we use for track spacing selection is to choose the minimum sensor range, corresponding to the minimum number of tracks.
The minimum number of tracks condition results in the most efficient mission.
There are usually multiple possible sensor range values that satisfy this condition.
Furthermore, once we know the minimum number of tracks that can cover the area, we also want to shift any possible coverage overlap to the tail of the $P_d$ curve, where the data quality is reduced.
We achieve this by selecting the minimum sensor range, out of the ones that give us the least number of tracks.
This value spreads out this overlap uniformly between all tracks, rather than having most of the overlap concentrated on the last pair of legs.

The sensor range upper boundary, $r_{eff}$, is adapted online based on sonar data and the updated $P_d$ curve.
This results in an online update of the track placement, following the new sensor range, $r_{adpt}$, and the remaining area that has not been covered, $W$.

This heuristic provides a quick computation of the reduced state space that considers only full coverage with paired tracks solutions.
In case the $P_d$ threshold violates the inequality in equation \ref{eq:sensor_range_nadir}, and we cannot cover the nadir gap with paired tracks, or if we need a more dense coverage, the problem should be modelled in a different way.
For example, track spacing within and between track pairs could be decoupled.
Or the placement of individual tracks could be approached as an instance of a set cover problem, e.g. see \cite{vazirani2013approximation}.

\section{Experiments}
We conducted three experiments to show the performance of the adaptive track spacing in typical operational situations - when the sensor range is overestimated, underestimated, and assumed optimal.

Figure \ref{fig:muscle} shows the assets we used to collect experimental data  - the MUSCLE, deployed from the coastal research vessel Leonardo.

We collected data using the adaptive track spacing algorithm on 18, 19 and 20 September 2019 in calm waters off the coast of Tellaro, Italy.
Following operational constraints and safety procedures, we selected a mission area with a width of 1212 metres and a length of 400 metres.

For all the experiments we ran, we had a control mission using predefined initial parameters with no adaptation enabled.
One important note is that the real maximum range of the SAS, $r_{max}$, was limited to 130 metres for all experiments --- no data was collected beyond this range.
We aligned the tracks with the short side of the survey area so we could observe a larger number of tracks, so the experiments did not follow a turn-minimisation strategy, as would have been the case in an actual mission \cite{yordanova2019coverage}.

\subsubsection{Experiment 1} Overestimating the sensor range ($r_{max}$ = 145) --- if the sensor range is overestimated, we expect that there will be gaps in the coverage. 
Setting the initial sensor range at $r_{max}$ = 145 for both the adaptive and predefined missions exceeded their ensonification limit ($r_{max}$ = 130 metres).
\subsubsection{Experiment 2} Minimum sensor range ($r_{max}$ = 120) --- if the sensor range is underestimated, we expect that there will be additional overlap that brings little performance improvement but leads to a considerable efficiency decrease. 
Based on equation \ref{eq:sensor_range_nadir}, we set the initial sensor range for the adaptive and the predefined missions to their minimum --- 120 metres.
\subsubsection{Experiment 3} Optimum sensor range ($r_{max}$ = 130) --- if we are confident in our estimate of the sensor range, and if the conditions do not change, we expect that there will be no need to readapt the tracks during the mission.
According to the MUSCLE operators, the `usual' sensor range of the SAS sonar under generally used operational parameters and sensor settings is 130 metres.

\section{RESULTS}
\label{results}

\begin{figure}[htbp]
	  \centerline{\includegraphics[width=1\linewidth]{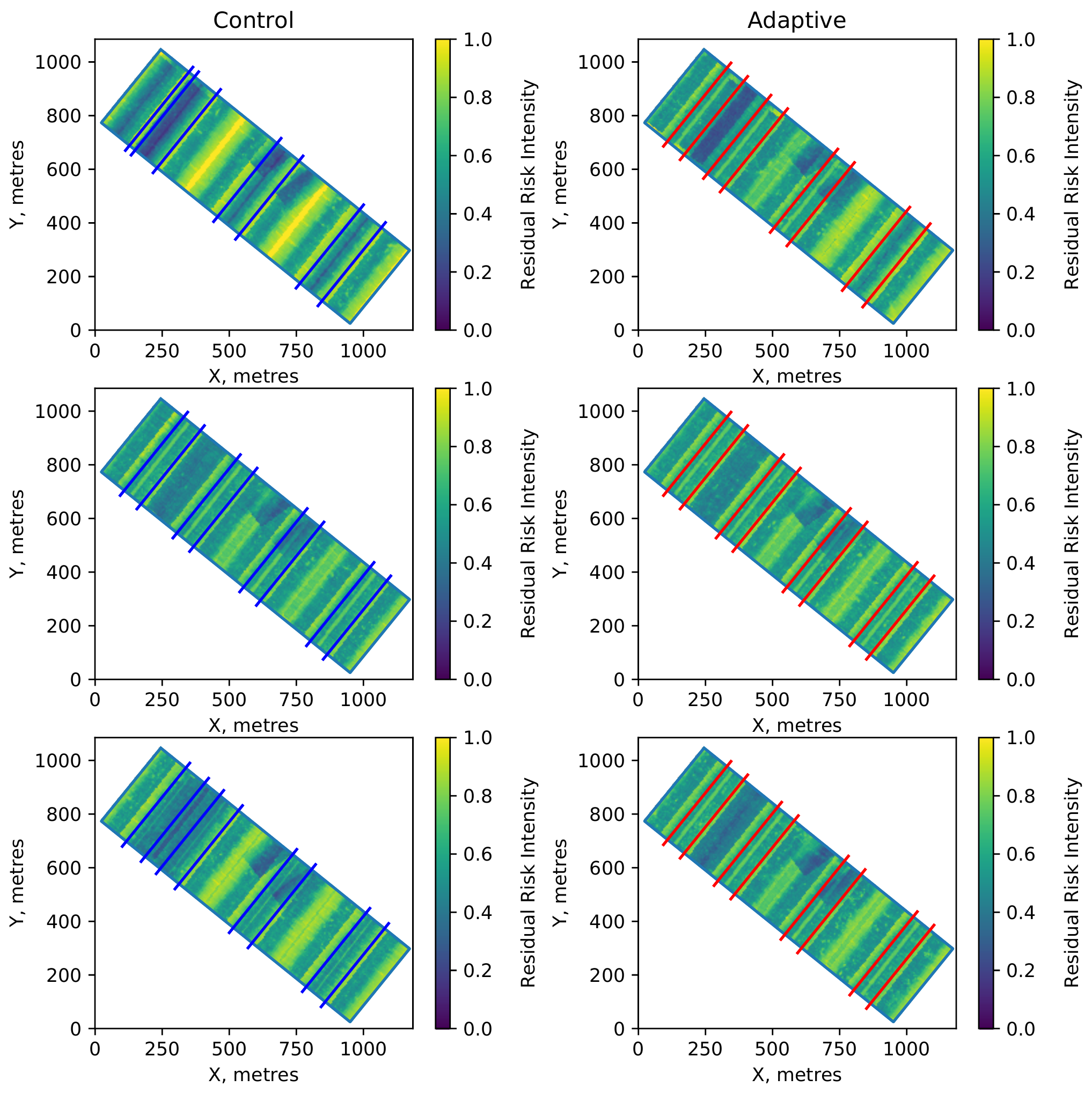}}
    \caption{Experimental data demonstrating the performance of the adaptive survey approach during three at-sea tests. The top row, shows Experiment 1 --- the effect on data collection when we overestimate the AUV sensor range. The middle plot is Experiment 2 --- underestimating the AUV sensor range. The bottom plot shows data from Experiment 3 --- aimed at showing the effect on data collection when we select the `optimal' AUV sensor range. The plots on the left display measured residual risk (the estimated probability of missing a target, see equation \ref{eq:RRM}) for a predefined (control) mission and on the right --- adaptive survey mission. Lower values, or dark blue, correspond to low residual risk, which is desirable. Higher values, in green, show high residual risk. The maximum risk is in yellow, corresponding to no sensor coverage. The location of the tracks is marked by lines, starting in the bottom right corner.}
	  \label{fig:all_exp_rrm_data}
	\end{figure}

Figure \ref{fig:all_exp_rrm_data} shows maps of the residual risk intensity together with the tracks' locations marked with coloured lines. 
The maps reflect the probability of missing a target if it were there, i.e. dark blue colour indicate good sensor coverage, whereas bright yellow values correspond to bad or no coverage.
Areas outside of the mission region are shown in white and we did not collect data there.
The mission area was benign, with uniform seabed, so the relatively high overall residual risk is likely due to a suboptimally trained model that generates the $P_d$ curve \cite{Gips2018a}.
We excluded the outermost perimeter area as there were missing data due to the processing of the map rather than to the data collection methodology.
This resulted in 16763 data points in total, where each data point corresponds to a 5x5 metres square area.

\subsection{Track spacing behaviour}
\subsubsection{Experiment 1 --- overestimating the sensor range}
	
The top row of Figure \ref{fig:all_exp_rrm_data} shows the data, from Experiment 1.
The control mission is on the left --- we placed the tracks at predefined sensor range of 145 metres, and did not allow any adaption to occur.
For the adaptive survey on the right, we initialised the sensor range at 145 metres but the Polygon adaptation (Algorithm \ref{algo:poly_adpt}) updated the sensor range $r_{adpt}$ at 138 metres in order to optimise the coverage overlap.
After the first pair of tracks, the AUV used the collected data to calculate a $P_d$ curve and update the admissible `good' data limit $r_{eff}$ to 130 metres, instead of the initial, overestimated 145 metres.
The Polygon adaptation took the width of the area, that was not yet covered after the first two tracks, and the new upper sensor range limit, and set $r_{adpt}$, autonomously, to 120 metres.
There was no further adaptation observed during the trial.

\subsubsection{Experiment 2 --- minimal sensor range}
The data we collected during Experiment 2 is shown in the middle row of Figure \ref{fig:all_exp_rrm_data}.
The left and the right plots, corresponding to a control and adaptive missions have the same track positions --- the adaptive survey did not trigger any change.
The initial $r_{max}$ value of 120 metres was updated to 130, after a new $P_d$ curve informed the vehicle that it could select a wider track spacing in the adaptive survey mission.
However, the Polygon adaptation resulted in keeping the tracks at 120 metres.

\subsubsection{Experiment 3 --- optimal sensor range}
In Experiment 3, we set the initial sensor range $r_{max}$ at 130 metres, for both the predefined and adaptive missions.
For the control mission, the sensor range stayed at 130 metres and the data we collected can be seen on Figure \ref{fig:all_exp_rrm_data}, in the left, bottom corner.
For the adaptive mission, the Polygon adaptation set the sensor range $r_{adpt}$ at 120 metres.
After a pair of tracks, the $P_d$ curve confirmed the initial expectation, that admissible data can be collected up to $r_{max}$ at 130 metres.
However, the Polygon adaptation optimised the spacing at 120 metres.
This decision was repeated after every pair of tracks.

\subsection{Coverage}
One of the main goals of the adaptive survey is to provide a full area coverage.
In Experiment 1, the gaps between track pairs in the control case were due to the fact that data was not collected beyond a range of 130 m --- the area that was not covered was 4.4\% (Table \ref{table:results_efficiency}) of the overall mission area.
When the vehicle detected the initial track spacing was too wide, it replanned based on the data, and reduced the area that was not covered to 0.18\%.
This difference can be seen on the top row of plots in Figure \ref{fig:all_exp_rrm_data}.
On the left, the control mission has visible yellow strips between the paired tracks, which means there is a maximum residual risk due to no coverage.
On the right, the adaptive survey mitigates this effect by shifting the yellow to light green - the residual risk is high, but within the predefined threshold.

\begin{table}
\centering
  \sisetup{
    table-figures-integer = 3                      ,
    table-figures-decimal = 0                      ,
    table-space-text-post = {~(\SI{99.99}{\percent})} ,
  }
  \def\x#1{~(\SI{#1}{\percent})}

\begin{tabular}{ c c c @{}S@{} c }
\Xhline{3\arrayrulewidth}
Experiment & Strategy & No Coverage & Num Tracks\\
\hline
\multirow{2}{*}{1 (r=145)}	&ctrl & 136 $m^{2}$ \x{4.40} & 7\\
				&adpt &  28 $m^{2}$ \x{0.18} & 8\\
\cline{2-4}
\multirow{2}{*}{2 (r=120)}	&ctrl &  0 $m^{2}$  \x{0}     & 8\\
				&adpt & 10 $m^{2}$ \x{<0.01} & 8\\
\cline{2-4}
\multirow{2}{*}{3 (r=130)}	&ctrl & 0 $m^{2}$  \x{0}     & 8\\
				&adpt & 7 $m^{2}$ \x{<0.01} & 8\\
\Xhline{3\arrayrulewidth}
\end{tabular}
\caption{Comparing mission efficiency metrics --- measured missed coverage and number of tracks for the three at-sea experiments: $r_{max} = 145$, $r_{max} = 120$ and $r_{max} = 130$. }
\label{table:results_efficiency}
\end{table}

The coverage gaps, in both the control and the adaptive missions in Experiments 2 and 3, are small enough (less than 0.2 \% of the total area, table \ref{table:results_efficiency}) that we can attribute them to errors in processing the data or projecting them to the residual risk maps, and ignore them.

\subsection{Number of tracks}
Experiment 1 resulted in a different overall path length for the adaptive and the control missions.
The adaptive survey required an additional track, which translates to 14\% increase in mission path length, or 400 metres.
In Experiments 2 and 3, both the adaptive and control missions, required the same number of tracks.

Adding or removing a track from the survey brings a dramatic change to the overall mission duration.
Due to the small margin (30 metres) in which we can shift the spacing of the MUSCLE's tracks (see the shaded area in Figure \ref{fig:range_to_spacing} - the difference between $r_{max}$ and the limit $3 \times r_{min}$), relative to the swath footprint width (100 metres per side) and the survey area width (1212 metres), this does not happen often.
The effect would be more prominent for a longer mission (larger area width) or larger sensor margin.
In addition, the operators in general know the limits of their system, making Experiment 3 the most likely scenario in a real setting.
In the case of uncertain environmental conditions, the more conservative setting of Experiment 2 may be more applicable.
Taken together, this means that often times, the number of tracks will be the same, regardless if we chose to adapt according to $r_{max}$ or $3 \times r_{min}$.
However, the difference in coverage quality matters, and this shows when we look at the data quality analysis.

\subsection{Data quality}
\begin{figure}[htbp]
	  \centerline{\includegraphics[width=1\linewidth]{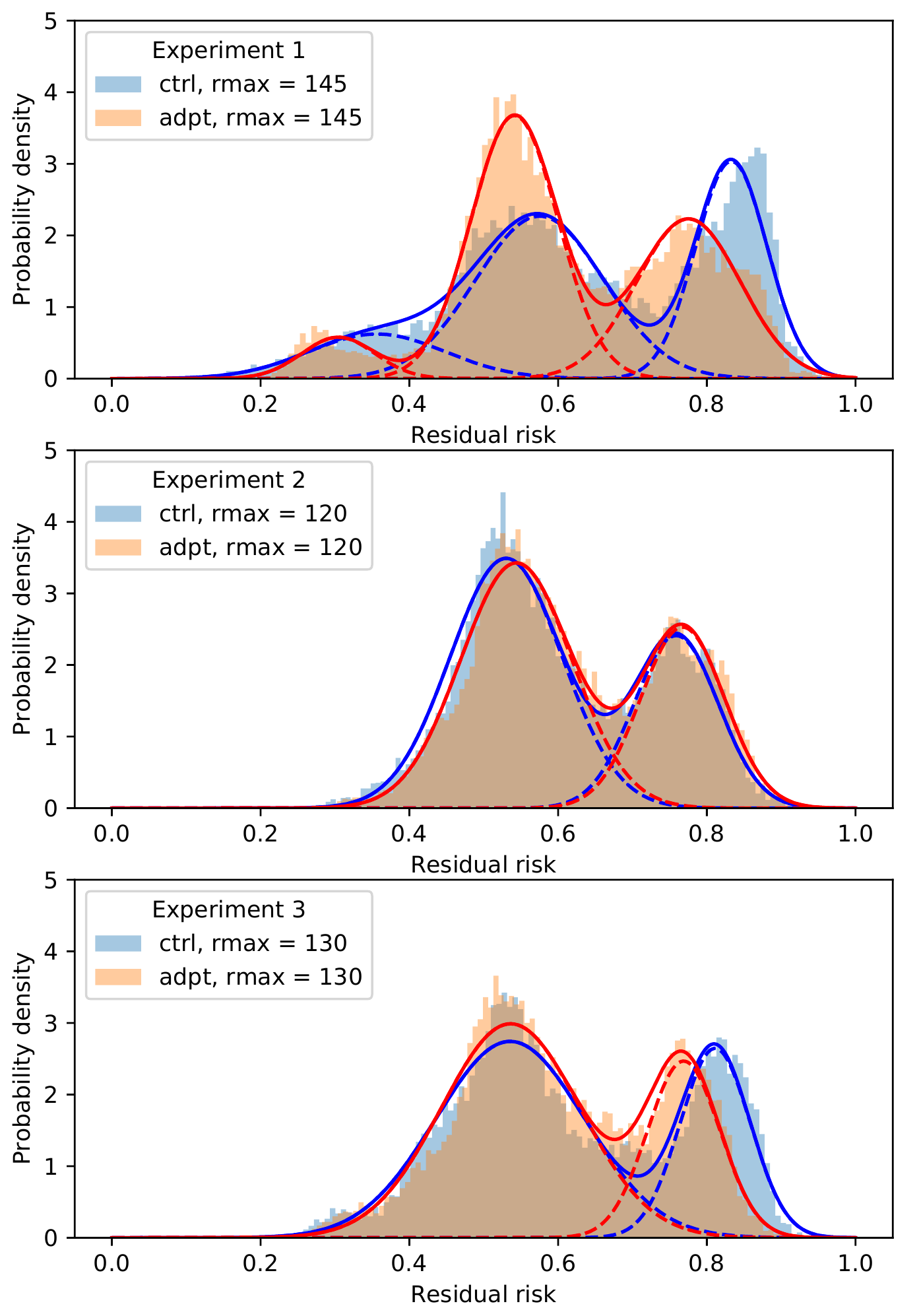}}
	  \caption{Gaussian mixture model (GMM) fits of histograms from the residual risk data from Figure \ref{fig:all_exp_rrm_data}. The data from experiments 1,2 and 3 are shown in the top, middle and bottom panel respectively. The blue histograms correspond to the preplanned missions (left plots in Figure \ref{fig:all_exp_rrm_data}). They are fitted with GMMs in blue continuous line and their components are in blue dashed lines. The orange histograms shows adaptive survey data (right column of plots on Figure \ref{fig:all_exp_rrm_data}). The red continuous lines are GMM fits and the dashed red lines give their components.}
	  \label{fig:hist}
	\end{figure}

To further compare the data, we plotted residual risk histograms of the control missions (in blue) and the adaptive survey missions (in orange) on Figure \ref{fig:hist}.
The top, middle and bottom plots correspond to Experiments 1,2 and 3 respectively.
The histograms do not show the data where the residual risk is 1 --- we treat this data as outliers and record it as `No Coverage' in Table \ref{table:results_efficiency}.

The dominant peak in all six histograms corresponds to the profile of the $P_d$ curve, shown in Figure \ref{fig:pd_profile}.
This is the residual risk we achieved with a single look of the sensor.
To the right and left of the dominant peak is the overlap.
The peak on the right, where the residual risk is high, is where we had coverage overlap of the tail of the $P_d$ curve.
This corresponds to light green in Figure \ref{fig:all_exp_rrm_data} and is what we aimed at improving with our adaptive track spacing algorithm.

In order to compare the histograms, we fitted Gaussian mixture models (GMM) to the data with three (experiment 1) or two (experiments 2 and 3) components.
The blue continuous lines are GMMs for the control missions histograms and the blue dotted lines show the individual components making up the mixture model.
The red lines are GMM fits for the adaptive survey histograms.

The summary statistics are given in Table \ref{table:results_rrm}.
The statistics of the full distributions are not informative for our analysis as it is hard to compare multimodal distributions, but their means are given in column `$\mu$ (f)', where (f) stands for `full'.
The rightmost components of the histograms show the effect that the adaptive track spacing method had on the data collection.
In Table \ref{table:results_rrm}, `(rc)' denotes `rightmost component', and we provide their means $\mu$, standard deviation $\sigma$ and number of data points, $n$.

\begin{table}
\centering
\begin{tabular}{ c c c c c c }
\Xhline{3\arrayrulewidth}
Experiment & Strategy & $\mu$ (f) & $\mu$ (rc) & $\sigma$ (rc) & n (rc)\\
\hline
\multirow{2}{*}{1 (r=145)}	&ctrl & 0.654 & 0.841 & 0.042 & 4992 \\ 
				&adpt & 0.617 & 0.796 & 0.059 & 4963 \\ 
\cline{2-6}
\multirow{2}{*}{2 (r=120)}	&ctrl & 0.609 & 0.759 & 0.057 & 5727 \\ 
				&adpt & 0.624 & 0.767 & 0.056 & 6003 \\ 
\cline{2-6}
\multirow{2}{*}{3 (r=130)}	&ctrl & 0.622 & 0.811 & 0.047 & 5235 \\ 
				&adpt & 0.606 & 0.769 & 0.049 & 5051 \\ 
\Xhline{3\arrayrulewidth}
\end{tabular}
\caption{Comparing data quality --- measured RRM mean, standard deviation and weight of the rightmost component of the Gaussian mixture model (GMM) for the three at-sea experiments: $r_{max} = 145$, $r_{max} = 120$ and $r_{max} = 130$ }
\label{table:results_rrm}
\end{table}

The adaptive survey does not result in reducing the overall residual risk.
The difference in mean between the control and adaptive surveys' full distributions in Experiment 3 (r=130, Table \ref{table:results_rrm}, $\mu$ (f)), where we had different track placements, is similar to that observed in Experiment 2 (r=120), where the track placement between the control and adaptive survey were the same.
The overall residual risk mean is not a reliable measure as the change between an adaptive survey and a control mission is within the variation of sonar ensonification.
We cannot claim that an improvement is due to track adaptation or lower overall noise as the difference is about 1.5\%.
We exclude Experiment 1 from this consideration since the adaptive survey required an additional track (Table \ref{table:results_efficiency}), and this led to a decrease in the mean residual risk compared with the control mission.

The adaptive survey reduces the residual risk where the `worst' data is collected by distributing the coverage overlap.
In the histogram of Experiment 3, the mean of the rightmost component of the adaptive survey is shifted compared with the control mission.
Employing the adaptive survey strategy during the mission resulted in 4.2\% reduction in residual risk for nearly 30\% of the `worst' data.

We refrain from interpreting the value of the absolute change in residual risk.
This is highly dependent on the mission requirements, such as the admissible $P_d$ threshold, the environmental conditions, as well on the size of the mission area and the length of the mission.
We do achieve a shift in residual risk and show that it mainly affects the areas with worst quality.
This is, however, a tradeoff --- increasing the data quality over previously ``bad'' areas, will result in a reduction of the ``best'' data quality.

\section{CONCLUSIONS}
We have developed, integrated and tested a coverage path planning algorithm for adaptive track spacing mine countermeasures survey mission using an autonomous underwater vehicle (AUV) equipped with a side-looking sonar.
The new adaptive survey combines both efficiency and data-related objectives in its cost function.

Adapting the track spacing of the AUV in real time has an effect on the quality of data collection.
Our experiments demonstrated that when the coverage overlap is shifted to the areas with lowest sensor quality, this results in improved data collection.

The next step is to make the coverage more uniform.
Instead of revisiting the area with the same survey vehicle that has a high-resolution sensor, it can be assisted by a smaller and more maneuverable vehicle, with a simpler sensor.
When the regions with lower coverage become apparent early on, a smaller vehicle can be tasked to revisit these hotspots in parallel.

Another line of research is developing a better understanding of the range-performance metric (e.g. $P_d$ curve) and use the whole profile for the track spacing optimisation, rather than rely on a threshold.

\addtolength{\textheight}{-15.3cm}   





\section*{ACKNOWLEDGMENT}

The authors would like to thank the whole MUSCLE engineering team, the CRV Leonardo crew and the CMRE autonomous naval mine countermeasures team (ANMCM) for their valuable assistance during the planning and execution of the experiments. We would like to thank our colleagues Simone Vasoli, Per Arne Sletner, Francesco Baralli and Tom Furfaro for all their efforts before, during and after the sea trials.


\bibliographystyle{ieeetr}
\bibliography{ral_20_cpp_auv_arxiv}

\end{document}